\pdfoutput=1

\documentclass[11pt]{article}

\usepackage[]{ACL2023}

\usepackage{times}
\usepackage{latexsym}
\usepackage[T1]{fontenc}
\usepackage{newtxtext,newtxmath}

\usepackage[utf8]{inputenc}

\usepackage{microtype}
\usepackage{mdframed}
\usepackage{graphicx}

\usepackage{inconsolata}
\newcommand\blfootnote[1]{%
  \begingroup
  \renewcommand\thefootnote{}\footnote{#1}%
  \addtocounter{footnote}{-1}%
  \endgroup
}
\usepackage{algorithm}
\usepackage{algorithmicx}
\usepackage{algpseudocode}
\usepackage{pifont}
\title{DEBATE: Devil's Advocate-Based Assessment and Text Evaluation}

\author{Alex G. Kim\textsuperscript{*} \\
  University of Chicago\\
  \texttt{alex.kim@chicagobooth.edu} \\\And
  Keonwoo Kim\textsuperscript{*} \\
  VRCREW \\
  \texttt{keonwookim@vrcrew.com} \\\And  
  Sangwon Yoon\textsuperscript{*}\\
  Ministry of Justice, Republic of Korea\\
  \texttt{sangwon38383@spo.go.kr}}

\begin{document}

\maketitle

\begin{abstract}\blfootnote{* indicates equal contribution.}As natural language generation (NLG) models have become prevalent, systematically assessing the quality of machine-generated texts has become increasingly important. 
Recent studies introduce LLM-based evaluators that operate as reference-free metrics, demonstrating their capability to adeptly handle novel tasks.
However, these models generally rely on a single-agent approach, which, we argue, introduces an inherent limit to their performance. 
This is because there exist biases in LLM agent's responses, including preferences for certain text structure or content.
In this work, we propose DEBATE, an NLG evaluation framework based on multi-agent scoring system augmented with a concept of \emph{Devil's Advocate}.
Within the framework, one agent is instructed to criticize other agents' arguments, potentially resolving the bias in LLM agent's answers.
DEBATE substantially outperforms the previous state-of-the-art methods in two meta-evaluation benchmarks in NLG evaluation, SummEval and TopicalChat.
We also show that the extensiveness of debates among agents and the persona of an agent can influence the performance of evaluators. 
\end{abstract}

\section{Introduction}
As the quantity of machine-generated texts have increased dramatically due to recent advances in generative large language models (LLMs) \citep{GPT3,ouyang2022training,gpt4,LM_survey}, robust NLG evaluation has become important \citep{sai2022survey, li2024leveraging}. Improving the performance of NLG evaluation largely hinges on reliable and consistent automatic evaluators. Traditional evaluation techniques like ROUGE \citep{Rouge} and BLEU \citep{saadany-orasan-2021-bleu} perform assessments at the token level, having limitations in reflecting contextualized semantics of sentences. Recent methods equipped with pre-trained language models incorporate contextual information in evaluation \citep{BERTScore, BARTScore, MoverScore, UniEval}. However, most existing methods exhibit a relatively low correlation with human ratings, and rely heavily on costly human-annotated data.

Recent research has introduced LLM-based methods for NLG evaluation \citep{selfcheckgpt, GPTScore, G-EVAL}. They usually require task-specific prompts and generate scores based on the probabilities of output tokens. On the other hand, these LLM-based methods, by predominantly utilizing single-agent systems without a mechanism for critiquing or post-processing LLM-generated responses, are particularly susceptible to implicit or induced biases \citep{koo2023benchmarking}. This limitation undermines both the stability and the effectiveness of their performance in evaluation tasks.
\begin{figure*}[!ht]
\setlength{\abovecaptionskip}{0.2cm}
\setlength{\belowcaptionskip}{-0.2cm}
\centering

\includegraphics[height=5.6cm]{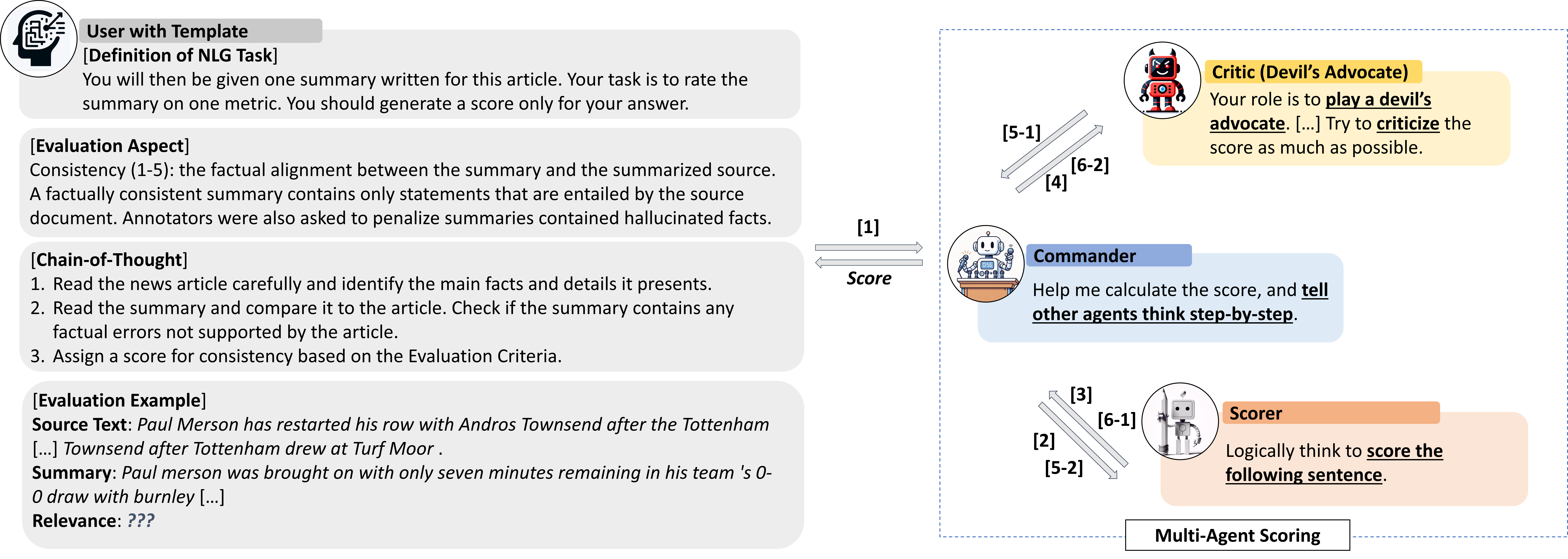}
\caption{Overall framework of DEBATE. Numbers around the arrows correspond to the steps in Appendix \ref{steps_debate}. The figure illustrates an example of deriving a consistency score in summarization task.}
\label{figure-1}
\end{figure*}



In this paper, we introduce DEBATE (Devil's Advocate-Based Assessment and Textual Evaluation), a simple yet effective multi-agent-based NLG evaluation framework that uses scores generated by LLMs. Based on the structure of \citet{macdougall1997devil}, we design a multi-agent scoring system where three agents, Commander, Scorer, and Critic, debate with each other. The Commander acts as the leader in debate while the Scorer calculates scores in response to given tasks. The Critic is given a role of \emph{Devil's Advocate} and provides constructive criticism on the Scorer's output, ensuring a more balanced evaluation process. We expect the framework to correct potential biases in NLG scoring and yield answers that align well with human-generated ones.

We show that debating strategies commonly used in social sciences, such as Devil's Advocate, have significant potential in LLM-based scoring systems. Despite its straightforward concept, experimental results indicate that DEBATE sets new state-of-the-art results on two NLG meta-evaluation benchmarks: summarization and dialogue generation.

\section{Method}

\begin{table*}[ht!]
\centering
\begin{footnotesize}
\begin{tabular}{llcccccccccc} 
\hline\hline
\multicolumn{2}{l}{\textbf{SummEval}}                                & \multicolumn{2}{c}{Average}                       & \multicolumn{2}{c}{Coherence}                     & \multicolumn{2}{c}{Consistency}                   & \multicolumn{2}{c}{Fluency}                       & \multicolumn{2}{c}{Relevance}                      \\ 
\cline{3-12}
               &                 & $\rho$ & $\tau$ & $\rho$ & $\tau$ & $\rho$ & $\tau$ & $\rho$ & $\tau$ & $\rho$ & $\tau$  \\ 
\hline
Others  
               & ROUGE-L\ding{61}         & 0.165                   & 0.128                   & 0.128                   & 0.099                   & 0.115                   & 0.092                   & 0.105                   & 0.084                   & 0.311                   & 0.237                    \\
               & BERTScore\ding{61}       & 0.225                   & 0.175                   & 0.284                   & 0.211                   & 0.110                   & 0.090                   & 0.193                   & 0.158                   & 0.312                   & 0.243                    \\
               & UniEval\ding{61}         & 0.474                   & 0.377                   & 0.575                   & 0.442                   & 0.446                   & 0.371                   & 0.449                   & 0.371                   & 0.426                   & 0.325                    \\
               & MOVERScore\ding{61}      & 0.191                   & 0.148                   & 0.159                   & 0.118                   & 0.157                   & 0.127                   & 0.129                   & 0.105                   & 0.318                   & 0.244                    \\
               & BARTScore\ding{61}   & 0.385                   & 0.305                   & 0.448                   & 0.342                   & 0.382                   & 0.315                   & 0.356                   & 0.292                   & 0.356                   & 0.273                    \\ 
               & GPTScore & 0.394                   & 0.344                       & 0.284                   & 0.242                       & 0.506                   & 0.480                       & 0.415                   & 0.390                       & 0.306                   & 0.265                        \\

Gemini Pro
               & G-Eval          & 0.290                   & 0.250                   & 0.350                   & 0.279                   & 0.403                   & 0.368                  & 0.158                   & 0.148                  & 0.248                   & 0.206                    \\
                & DEBATE          & 0.421                   & 0.356                   & 0.467                   & 0.365                   & 0.521                   & 0.501                   & 0.311                   & 0.302                   & 0.386                   & 0.256                    \\

GPT-3.5 
               & G-Eval          & 0.387                   & 0.320                   & 0.522                   & 0.406                   & 0.396                   & 0.331                  & 0.277                   & 0.267                  & 0.354                   & 0.277                    \\
               & MultiAgent      & 0.350                   & 0.309                   & 0.407                   & 0.348                   & 0.354                   & 0.325                   & 0.241                   & 0.223                   & 0.398                   & 0.341                   \\
                & DEBATE          & 0.472                   & 0.447                   & 0.565                   & 0.521                   & 0.516                   & 0.492                   & 0.397                   & 0.386                   & 0.410                   & 0.389                    \\

GPT-4   
               & G-Eval          & 0.533                   & 0.450                   & 0.522                   & 0.420                   & 0.600                  & 0.553                   & 0.495                   & 0.416                   & 0.517                   & 0.412                  \\
               & ChatEval      & 0.528                   & 0.458                   & 0.477                   & 0.412                   & 0.595                   & 0.516                   & 0.501                   & 0.432                   & 0.539                   & 0.472                    \\

               & MultiAgent      & 0.507                   & 0.459                   & 0.486                   & 0.410                   & 0.628                   & 0.596                   & 0.410                   & 0.393                   & 0.505                   & 0.436                    \\

                & DEBATE          & \textbf{0.597}                   & \textbf{0.575}                   & \textbf{0.610}                   & \textbf{0.588}                   & \textbf{0.643}                   & \textbf{0.614}                   & \textbf{0.542}                  & \textbf{0.515}                   & \textbf{0.593}                   & \textbf{0.582}                    \\

\hline\hline
              \multicolumn{2}{l}{\textbf{Topical-Chat}}                 & \multicolumn{2}{c}{Average}                       & \multicolumn{2}{c}{Naturalness}                   & \multicolumn{2}{c}{Coherence}                     & \multicolumn{2}{c}{Engagingness}                  & \multicolumn{2}{c}{Groundedness}                   \\ 
\cline{3-12}
               &                 & $r$ & $\rho$ & $r$ & $\rho$ & $r$ & $\rho$ & $r$ & $\rho$ & $r$ & $\rho$  \\ 
\hline

Others   
& ROUGE-L\ding{61}         & 0.243                   & 0.244                   & 0.176                   & 0.146                   & 0.193                   & 0.203                   & 0.295                   & 0.300                   & 0.310                   & 0.327                    \\
               & BERTScore\ding{61}       & 0.262                   & 0.273                   & 0.226                   & 0.209                   & 0.214                   & 0.233                   & 0.317                   & 0.335                   & 0.291                   & 0.317                    \\
               & UniEval\ding{61}         & 0.552                   & 0.417                   & 0.455                   & 0.330                   & 0.602                   & 0.455                   & 0.573                   & 0.430                   &0.577                   & 0.453                    \\
               & MOVERScore      & 0.222                   & 0.238                   & 0.169                   & 0.170                   & 0.247                   & 0.259                   & 0.275                   & 0.269                   & 0.198                   & 0.147                    \\
               & BARTScore   & 0.293                   & 0.276                   & 0.287                   & 0.266                   & 0.251                   & 0.225                   & 0.411                   & 0.406                   & 0.226                   & 0.205                    \\
                & GPTScore & 0.367                   & 0.345                       & 0.353                   & 0.289                       & 0.279                   & 0.223                       & 0.472                   & 0.451                       & 0.362                   & 0.417                        \\

Gemini Pro
               & G-Eval          & 0.353                   & 0.324                   & 0.266                   & 0.235                   & -0.015                   & 0.006                  & 0.508                   & 0.438                  & 0.651                   & 0.616                    \\
                & DEBATE          & 0.450                   & 0.418                   & 0.402                   & 0.412                   & 0.191                   & 0.113                   & 0.541                   & 0.516                   & 0.663                   & 0.631                    \\
            
GPT-3.5  
               & G-Eval          & 0.399                   & 0.394                  & 0.348                   & 0.362                   & 0.510          & 0.500                   
               & 0.355             & 0.361                   
               & 0.385                   & 0.353                    \\

               & MultiAgent      & 0.368                   & 0.363                   & 0.237                  & 0.231                   & 0.295                   & 0.302                   & 0.540                   & 0.523                   & 0.398                  & 0.397                    \\
                & DEBATE          & 0.499                   & 0.452                   & 0.502                   & 0.469                   & 0.485                   & 0.430                   & 0.556                   & 0.480                   & 0.452                   & 0.428                    \\

GPT-4   
               & G-Eval          & 0.616                   & 0.623                   & 0.632                   & 0.609                   & 0.614                   & 0.619                   & 0.547                   & 0.579                   & 0.670                   & 0.684                    \\
               & ChatEval  & 0.675                   & 0.653                   & 0.620                   & 0.604                   & 0.657                   & 0.641                   & 0.687                   & 0.676                   & 0.737                   & 0.692                    \\
               
               & MultiAgent      & 0.656                   & 0.649                   & 0.590                   & 0.570                   & 0.665                   & 0.667                   & 0.709                   & \textbf{0.729}                   & 0.659                   & 0.630                    \\
                & DEBATE          & \textbf{0.735}                   & \textbf{0.729}                   & \textbf{0.699}                   & \textbf{0.683}                   & \textbf{0.735}                   & \textbf{0.727}                   & \textbf{0.761}                   & 0.705                   & \textbf{0.745}                   & \textbf{0.800}                    \\

\hline\hline
\end{tabular}
\caption{\label{main_results}
Table shows summary-level Spearman ($\rho$) and Kendall-Tau ($\tau$) correlations for SummEval, alongside Pearson ($r$) and Spearman ($\rho$) correlations for Topical-Chat, between machine and human evaluations. \ding{61}: results from \citet{G-EVAL}; all other results are reproduced by the authors. See Appendix \ref{detail_settings} for experiment details.
}
\end{footnotesize}
\end{table*}

\label{method}
\paragraph{Overview} 
DEBATE, an LLM-based multi-agent NLG evaluator, adopts the experimental design from \citet{macdougall1997devil} and includes three agents: Commander (leader), Scorer, and Critic. As in Figure \ref{figure-1}, DEBATE has three primary components: (1) a predefined template that outlines the NLG task (e.g., summarization) and the evaluation aspect (e.g., coherence), (2) a multi-agent scoring framework that embeds a debate among agents, and (3) a prompt of Devil's Advocate to criticize the Scorer. See Appendix~\ref{detail_settings} for details.


\paragraph{Predefined template}
We refer to the prompts used in ~\citet{G-EVAL} as a basis for the task description and aspect definition. Since DEBATE is designed to facilitate logical thinking among agents through consistent debate standards, we apply a zero-shot chain-of-thought~\citep{wei2022chain}.

\paragraph{Multi-agent scoring framework}

In the multi-agent scoring framework outlined in Algorithm \ref{pseudo_code}, the three agents play distinct roles. 
The Commander, following \citet{macdougall1997devil}, promotes dialogues between the Critic and Scorer by facilitating the transfer of information from previous debates. As LLM agents lack the ability to recall past conversations, it is necessary to include the Commander in our framework.; thus, the Commander acts as collective memory. The Scorer assesses texts based on specific instructions. The Critic engages in a debate with the Scorer and offers constructive criticism, playing the role of a Devil's Advocate.


\paragraph{Devil's Advocate}
Applying the concept of Devil's Advocate, the prompt for the Critic is: 

\begin{mdframed}
\emph{Your role is to play a Devil's Advocate. Your logic has to be step-by-step. Critically review the score provided and assess whether the score is accurate. If you don't think that the score is accurate, criticize the score. Try to criticize the score as much as possible.} 
\end{mdframed}

\begin{algorithm}[!ht]
\caption{Multi-Agent Scoring Framework}
\begin{algorithmic}[1]
\Require NLG task (T) and aspects (A)
\State Initialize $n$, max iterations
\State Define agents: Commander ($C$), Scorer ($S$), Critic ($Cr$)
\State $P \gets C(T, A)$ \Comment{Formulate prompts}
\State $Score \gets S(P)$ \Comment{Calculate score}
\For{$i=1$ to $n$}
    \State $C.send(P, Score, to=Cr)$  \Comment{Forward}
    \State $Feedback \gets Cr(Score)$ \Comment{Validate}
    \If{`NO ISSUE' in $Feedback$}
        \State \textbf{break}
    \Else
        \State $C.send(Feedback, to=S)$
        \State $Score \gets S(Feedback)$
    \EndIf
\EndFor
\State Determine final $Score$
\Ensure Final score for task
\end{algorithmic}
\label{pseudo_code}
\end{algorithm}

\section{Experiment}
\label{experiment}



\paragraph{Dataset}
We evaluate DEBATE on two meta-evaluation benchmark datasets, focusing on its correlation with human assessments. (i) SummEval \citep{SummEval} is a summarization evaluation dataset, providing human-assigned ratings across four key dimensions: \emph{fluency}, \textit{coherence}, \textit{consistency}, and \textit{relevance}. (ii) Topical-Chat \citep{Topical-Chat} is a knowledge-grounded human-to-human conversation dataset, and we refer \citet{UniEval} to evaluate four dimensions: \textit{naturalness}, \textit{coherence}, \textit{engagingness}, and \textit{groundedness}.

\paragraph{Baselines}
We extensively evaluate the performance of DEBATE with eight baselines, including a traditional evaluator, ROUGE-L \citep{Rouge}; the pretrained language model-based evaluators, BERTScore \citep{BERTScore}, MoverScore \citep{MoverScore}, BARTScore \citep{BARTScore}, and UniEval \citep{UniEval}; the recent LLM-based evaluators, GPTScore \citep{GPTScore}, G-Eval \citep{G-EVAL}, and ChatEval \citep{chan2023chateval}. We also include MultiAgent, a framework similar to DEBATE but with the Critic assigned a neutral debating role, denoted as \emph{Plain}.
\begin{figure*}[hbt!]

\centering
\includegraphics[width=160mm]{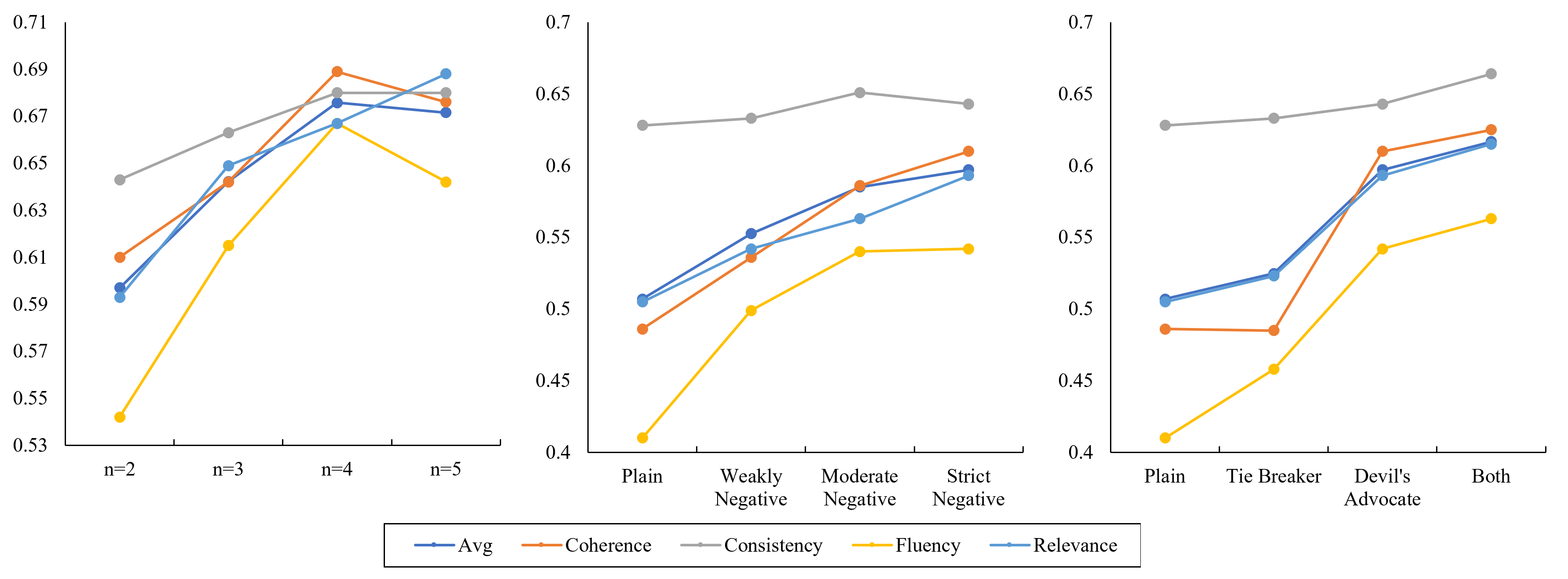}
\caption{All experimental results shown in this figure are obtained using DEBATE with GPT-4 on the SummEval dataset, illustrating the effect of $n$ (the number of maximum iterations) (left), agent persona (middle), and debating strategies (right) on model performance.`n' refers to the nubmer of debate iterations among multi-agents, and `Both' refers to adopting tie-breaker and Devil's Advocate simultaneously. See Appendix \ref{system_messages} for experiment details.}
\label{figure2}
\end{figure*}
\paragraph{Main results} 
Table \ref{main_results} shows the evaluation results on two meta-evaluation benchmarks. First, we find that MultiAgent is at least at par with the previous state-of-the-art framework, G-Eval, showing the potential of introducing multiple LLM agents in NLG evaluation. Notably, we observe that DEBATE, even though having no structural differences from MultiAgent, significantly outperforms all existing baselines on almost every benchmark. In particular, DEBATE substantially outperforms G-Eval, achieving 6.4\%p higher $\rho$ and 12.5\%p higher $\tau$ than G-Eval in SummEval; 11.9\%p higher $r$ and 10.6\%p higher $\rho$ than G-Eval in Topical-Chat. 

DEBATE also outperforms ChatEval, which is also based on multi-agent framework. By leveraging Devil's Advocate, DEBATE achieves new state-of-the-art results. These results imply that using opposing debaters to avert groupthink, \textit{i.e}., Devil's Advocate, unlocks the potential of multi-agents to serve as more effective evaluators. In addition, we demonstrate consistent efficiency of DEBATE using a different LLM family, Gemini Pro 1.5, which suggests that our results are not likely to be confined to OpenAI's LLM family only. Further, we include dialogues generated by DEBATE in Appendix \ref{Qualitative_results} and illustrate how the model manages to resolve the bias in the initial evaluation.





\paragraph{Number of iterations}
We conduct an ablation experiment to study the effect of the number of iterations on DEBATE's performance. In Figure \ref{figure2} (left), we find that, the model performance improves with more iterations on average. However, the performance reaches its plateau at $n=4$ and slightly declines at $n=5$. This indicates that sufficient debates may be generally helpful in improving the performance yet there may exist an optimal $n$.

\paragraph{Persona of the Critic}
In DEBATE, we assume a strong form of Devil's Advocate by instructing the model to criticize the other agent's answer as much as possible. 
In this experiment, we modify the persona of the Critic by changing the degree of criticality: `strictly negative' (DEBATE), `moderately negative', `weakly negative', and  `plain' (MultiAgent). Prompts for each persona are in Appendix~\ref{system_messages}. 
Figure \ref{figure2} (middle) shows that heightened criticality improves performance, implying that appropriate personas reinforce human-LLM alignment.

\paragraph{Debating style} We study the effect of applying another debating strategy on NLG evaluation. Even when a debate remains inconclusive, instead of taking the last response of the Scorer, we introduce Tie-Breaker, detailed in Appendix~\ref{detail_settings}.3, that collectively evaluates the debate history. As shown in Figure \ref{figure2} (right), Tie-Breaker achieves a better performance than MultiAgent (plain) yet a worse performance than DEBATE. However, we achieve the best performance when DEBATE is coupled with Tie-Breaker, indicating that combining different debating skills may improve the performance.

\section{Related work}

\paragraph{LLM-based evaluation}
LLMs have been extensively used as zero-shot evaluators for text evaluation \citep{LLM_evaluation, selfcheckgpt, prometheus,chatGPT_preliminary_study,G-EVAL_LLM_EVALUATION_ANALYSIS,LLM_evaluation_bad, LLM-Eval}. 
Most research presents a straightforward method to assess the quality of a text by providing task-specific instructions. Specifically, GPTScore \citep{GPTScore} assumes that higher-quality instructions and context correlate with higher probabilities based on GPT-3. G-Eval \citep{G-EVAL} implements an auto chain-of-thoughts approach. Unlike existing methods that predominantly use a single-agent prompting for NLG evaluation, our framework uses a multi-agent structure to generate more reasonable scores from debates.

\paragraph{Multi-agents with LLMs}
As the complexity and scope of tasks implemented by LLMs \citep{bang2023multitask,jiao2023chatgpt} increase, recent studies employing multi-agents have become more prevalent \citep{wang2023survey, xi2023rise, autogen, wu2023empirical}. Previous research has primarily used multi-agents for tasks such as divergent thinking \citep{dot_multi_agent}, as well as for improving factuality and reasoning \citep{du2023improving}. 

More recently, \citet{chan2023chateval} develop an evaluation framework, ChatEval, which also incorporates multiple LLM agents. DEBATE and ChatEval are distinct in several ways. First, their evaluation algorithms differ significantly. In ChatEval, each LLM agent independently generates scores through simultaneous discussions, and the final score is an average of these scores. In contrast, DEBATE employs a single scoring agent (Scorer) with a Devil’s Advocate providing critical feedback. If they fail to agree after several iterations, the Scorer’s final score is used. This methodology allows direct monitoring of the Scorer’s evaluation changes. Secondly, DEBATE relies on LLM agents' personas while ChatEval uses six roles without detailing the best-performing combinations. DEBATE, based on communication theories, explicitly assigns Devil’s Advocate, Scorer, and Commander roles, focusing on the impact of a strong Devil’s Advocate persona, ensuring consistent performance.

\paragraph{Debating strategies}
In group-level decision-making, groupthink has posed major threat to decision quality \citep{janis2008groupthink}. 
\citet{macdougall1997devil} show that assigning a role of `critic' helps avoid groupthinking and provides higher-quality outcomes. Such critic, labeled as Devil's Advocates, always provides an opposing view to others' opinions. We expect applying the concept of Devil's Advocate in multi-agent debate to \textit{unbias the bias} contained in single-agent answers.

\section{Conclusion}
In this paper, we propose DEBATE, the first state-of-the-art, multi-agent NLG evaluator with Devil's Advocate. We demonstrate the potential of incorporating debating strategies from social science into multi-agent scoring systems. We expect further exploration of the effect of diverse debating strategies in NLG evaluation.





\section*{Limitations}
While our proposed meta-evaluation method, DEBATE, demonstrates superior performance on benchmark datasets, surpassing other methods that use a single-agent approach, it is important to note that the implementation of a multi-agent system inherently incurs higher costs. Therefore, when considering the applicability of DEBATE, processing costs must always be taken into account. Also, we have not been able to test with LLM that have a relatively smaller number of parameters, such as Llama2~\citep{llama2}. Table~\ref{main_results} shows that DEBATE, when using GPT-3.5, performs worse than its counterparts utilizing GPT-4, indicating that its performance is significantly influenced by the inherent capabilities of the LLM itself. Thus, employing multi-agents in DEBATE with small-sized LLM might not yield optimal performance. Lastly, while Appendix~\ref{Qualitative_results} presents a qualitative analysis of the conversations between the agents in DEBATE, a quantitative evaluation of the various rational responses within the conversation has not been conducted. We plan to develop future evaluation methods to address this gap.

\section*{Ethics statement}
Our work introduces a novel evaluator that assesses the quality of generated texts, showing high correlation scores with human ratings. Nevertheless, we must consider the potential negative societal impacts that could arise from its inappropriate use. These concerns encompass ethical challenges such as automatically generating fake news, publishing machine-generated texts that receive high scores from our evaluator, or utilizing our technology for fraudulent activities. These implications highlight the need for responsible use and governance of such technologies, underscoring the importance of ethical considerations in the development and deployment of natural language processing tools.

\section*{Acknowledgements}
We appreciate insightful comments from Hyowon Cho, Joonwon Jang, MinJin Jeon, Jaehee Kim, Sangyeop Kim, Yukyung Lee, Maximilian Muhn, and Valeri Nikolaev. Yoon appreciates support from GenIP Co.,Ltd. and Blynx Co.,Ltd.

\bibliography{anthology,custom, references}
\bibliographystyle{acl_natbib_orig}

\appendix
\label{sec:appendix}

\section{Issues on replicating the performance of G-Eval}
We fail to replicate the performance of the G-Eval \cite{G-EVAL}. \citet{chiang-lee-2023-closer} also note the same issue. The reasons identified are as follows:

\begin{enumerate}
    \setlength\itemsep{-0.30em}
    \item The official implementation details for G-Eval\footnote{https://github.com/nlpyang/geval} are insufficient. Many exceptions arise in score parsing yet it is not clear how the original authors dealt with these issues. Further, the hyperparameter choice in the repository does not match the descriptions in the paper. For example, the paper states that the temperature is set to 1, but in the repository, it is set to 2. 
    \item The prompts used in Topical-Chat, except for the prompt used for evaluating engagingness, are not disclosed. 
    
    
    \item Since G-Eval repository has been made public, GPT-3.5 and GPT-4 models have undergone several internal changes. Such updates can also have a negative impact on model performance.\footnote{https://community.openai.com/t/gpt-4-is-getting-worse-and-worse-every-single-update/508470}
\end{enumerate}

\begin{table*}[ht!]
\centering
\begin{footnotesize}
\begin{tabular}{llcccccccccc} 
\hline\hline
\multicolumn{2}{l}{\textbf{SummEval}}                                & \multicolumn{2}{c}{Coherence}                       & \multicolumn{2}{c}{Consistency}                     & \multicolumn{2}{c}{Fluency}                   & \multicolumn{2}{c}{Relevance}\\ 
\cline{3-10}
               &                 & $Human$ & $DEBATE$ & $Human$ & $DEBATE$ & $Human$ & $DEBATE$ & $Human$ & $DEBATE$\\ 
\hline
Score  
               & 1         & 2.37                   & 5.23                   & 2.19                   & 6.23                   & 0.62                   & 7.71                  & 0.69                   & 6.22                 \\
               & 2       & 20.36                   & 18.23                 & 5.12                  & 4.34                   & 2.44                   & 2.32                   & 7.50                   & 6.27 \\
               & 3         & 26.36                   & 23.15                  & 2.31                   & 2.22                   & 5.12                   & 0.51                   & 22.61                   & 16.11 \\
               & 4      & 36.16                  & 34.15                  & 3.69                   & 2.30                   & 11.12                   & 1.23                   & 52.28                   & 45.23           \\
               & 5   & 14.74                   & 19.24                  & 86.70                   & 84.91                   & 80.70                   & 88.22                   & 16.93                   & 26.17 \\ 
\hline\hline
              \multicolumn{2}{l}{\textbf{Topical-Chat}}         &        \multicolumn{2}{c}{Naturalness}                   & \multicolumn{2}{c}{Coherence}                     & \multicolumn{2}{c}{Engagingness}                  \\ 
\cline{3-10}
               &                 & $Human$ & $DEBATE$ & $Human$ & $DEBATE$ & $Human$ & $DEBATE$\\ 
\hline
Score   
& 1         & 14.68                  & 19.43                   & 17.73                   & 21.00                  & 20.22                  & 35.42  \\
               & 2       & 40.72 & 26.55                   & 39.89                   & 30.15                   & 43.49                   & 31.13 \\
               & 3         & 44.60                   & 54.02                   & 42.38                   & 48.85                   & 36.29                   & 33.45  \\
              
\hline\hline
\end{tabular}
\caption{\label{count}Score distribution of Human and DEBATE for SummEval and Topical-Chat. We omit the 'groundedness' metric from these tables as it operates on a different scale. The numbers in the table are presented on a percentage scale (\%).
}
\end{footnotesize}
\end{table*}

\begin{table*}[h]
\centering
\begin{footnotesize}
\begin{tabular}{llcccccccccc} 
\hline\hline
\multicolumn{2}{l}{\textbf{SummEval}}                                & \multicolumn{2}{c}{Average}                       & \multicolumn{2}{c}{Coherence}                     & \multicolumn{2}{c}{Consistency}                   & \multicolumn{2}{c}{Fluency}                       & \multicolumn{2}{c}{Relevance}                      \\ 
\cline{3-12}
               &                 & $\rho$ & $\tau$ & $\rho$ & $\tau$ & $\rho$ & $\tau$ & $\rho$ & $\tau$ & $\rho$ & $\tau$  \\ 
\hline
DEBATE  
               & \emph{Sophisticated}         & 0.601                   & 0.573                   & 0.605                   & 0.592                   & 0.655                   & 0.610                   & 0.556                   & 0.521                   & 0.588                   & 0.569                    \\
  
               & \emph{Original}         & 0.597                   & 0.575                   & 0.610                   & 0.588                   & 0.643                   & 0.614                   & 0.542                   & 0.515                   & 0.593                   & 0.582                    \\
\hline\hline
\end{tabular}
\caption{\label{prompt_complexity}
Table shows summary-level Spearman ($\rho$) and Kendall-Tau ($\tau$) correlations for SummEval between machine and human evaluations on two kinds of prompt. \emph{Sophisticated} refers to more complex prompt than the \emph{Original} prompt originally used in DEBATE.
}
\end{footnotesize}
\end{table*}

\section{Score distribution of DEBATE}
Table \ref{count} illustrate the distribution of responses from humans and DEBATE. Interestingly, humans are likely to give extreme scores of 5 (or 3). DEBATE also follows a similar pattern. For Topical-Chat, DEBATE is more likely to give a score of 3 than humans. There could be two reasons for this: (i) the actual texts are of very good quality (ii) humans and DEBATE are both more likely to produce extreme scores than single-agent LLMs. In either scenario, this result also speaks to the same conclusion that DEBATE closely aligns with human evaluations.

\section{Prompt type of Devil's Advocate}
We conduct experiments varying the complexity of the Devil's Advocate prompt used in DEBATE. We evaluate the performance on SummEval using both the prompt from the existing setting (Original) and a more sophisticated and complex prompt (Sophisticated). As shown in Table~\ref{prompt_complexity}, the complexity of the prompt does not result in significant performance differences, indicating that DEBATE is robust to the prompts used for the Devil's Advocate. The Sophisticated prompt is as follows:

\begin{mdframed}
\emph{Your role is to play a Devil’s Advocate. Devil’s Advocate intentionally opposes the other’s opinion by providing constructive criticism. Refer to the scoring rubric and the score provided by the Scorer. Your actions may include, but are not confined to, the following: Try to criticize the score and the rationale step-by-step. Ask incisive questions and doubt any assumptions made by the Scorer. If the Scorer made any subjective judgment, try to criticize it as much as possible. If you can, come up with any hypothetical and counterfactual situation. Print out “NO ISSUES” only when you are absolutely sure that the Scorer’s answer is objective and accurate.} 
\end{mdframed}

\section{Further explanation in DEBATE}
\subsection{DEBATE}
\label{steps_debate}
This section explains the operational principles of DEBATE as covered in Algorithm \ref{pseudo_code} of Section \ref{method} using six specific steps. DEBATE consists of three interactive LLM agents (Commander, Scorer, and Critic), each with a specific role. The agents are capable of transmitting and receiving information amongst themselves. 

The six steps implemented in DEBATE are as follows:

\begin{enumerate}
    \setlength\itemsep{-0.30em}
    \item The process begins when a user inputs an NLG evaluation task and desired aspects as prompts into the system.
    \item The Commander provides prompts related to the task and aspect to the Scorer, requesting a score and its rationale.
    \item The Scorer then calculates and presents a corresponding score to the Commander.
    \item The Commander passes the response from the Scorer, along with the score and rationale, to the Critic for validation and feedback.
    \item Using the Critic's feedback, the Commander instructs the Scorer to adjust the score.
    \item The revised score is then subject to further validation by the Critic. When additional feedback arises, the cycle recommences from step 2, continuing in a loop until the Critic issues a `NO ISSUE' statement.
\end{enumerate}
The process iterates from step 2 to 6 until the number of iterations reaches \emph{n}, a predefined hyperparameter. The final score is determined by the Scorer's latest response if the Critic declares `NO ISSUE'; otherwise, by the score from the \emph{n}-th iteration. Although possibly suboptimal, the score from the \emph{n}-th iteration strikes a computational balance: increasing $n$ can improve outcomes by avoiding inconclusive debates but raises processing costs.

\section{Details of experiment}
\label{detail_settings}
\subsection{LLM settings}
For DEBATE, we use GPT-4 and GPT-3.5 provided by OpenAI in the form of API. Specifically, we perform the experiments with \texttt{gpt-4-1106-preview} and \texttt{gpt-3.5-turbo-1106}. When this study was conducted, the cost for processing input tokens with \texttt{gpt-4-1106-preview} was \$0.01 per 1,000 tokens, while generating output tokens was priced at \$0.03 per 1,000 tokens. For \texttt{gpt-3.5-turbo-1106}, the cost for processing input tokens was \$0.001 per 1,000 tokens, and for generating output tokens, it was \$0.002 per 1,000 tokens.

We uniformly apply the same hyperparameters across all GPT versions: a temperature of 0, both frequency and presence penalties set to 0, and a Top-p value of 1 for the cumulative probability cutoff used in nucleus sampling. Since we set the temperature hyperparameter as 0, we repeat the experiment once. As due diligence check, we randomly sample five observations and repeat the experiment 20 times to ensure that the results do not vary.


\subsection{Multi-agent implementation}
\citet{autogen}\footnote{https://github.com/microsoft/autogen} serve as a basis for MultiAgent implementation. AutoGen is a framework designed to facilitate the development of multi-agent LLM applications. We implement the interactions among the Commander, the Scorer, and the Critic using \texttt{MultiAgentsDebate} class, and each agent is implemented using \texttt{AssistantAgent} object.

\subsection{Tie-breaker}
We also introduce Tie-breaker, a novel debating strategy inspired by \citet{moderator} as discussed in Section~\ref{experiment}. 
Tie-breaker is the fourth agent that makes the final decision when the Scorer and the Critic fail to reach an agreement. When tie-breaker is implemented, we execute the six steps as in Appendix~\ref{steps_debate}. However, a difference arises when the Critic does not declare `NO ISSUE' within $n$ iterations. Instead of naïvely using the score from the n-th iteration, Tie-Breaker decides the score by collectively evaluating the dialogue history of the LLM agents up to that point. The prompt used for Tie-breaker is in Appendix~\ref{prompt_example}.

\subsection{Prompts and system message}
Examples of prompts and system messages used by the agents can be found in Appendix \ref{prompt_example}. The system messages of Plain, Weakly Negative, and Moderate Negative, as demonstrated in the examples, are used for the Critic's system messages in the "Persona of the Critic" experiment conducted in Section \ref{experiment}. We design the prompts to progressively increase the intensity of criticism provided by the Critic.

\subsection{Dataset}
\paragraph{SummEval} is a benchmark by Yale LILY Lab and Salesforce Research for evaluating summarization models on the English CNN/DailyMail dataset. It consists of a total of 1600 samples, which are 100 different source texts, each with 16 different versions of summaries. We calculate Spearman and Kendall tau correlation scores for each source text and then average them at the summary level.

\paragraph{Topical-Chat} is a benchmark for evaluating English dialogue response generation systems using knowledge. It comprises a total of 360 samples, which are 60 source texts, each with 6 facts and responses. We calculate Pearson and Spearman correlation scores for each source text and then average them at the text level.
\section{Qualitative analysis}
\label{Qualitative_results}
We present two illustrative examples to qualitatively analyze and evaluate the results of G-Eval, MultiAgent, and DEBATE, along with corresponding human scores. First, Appendix \ref{summeval_ex} presents the inference outcomes of each methodology on a Fluency aspect example from the SummEval dataset. Following this, Appendix \ref{topical_ex} shows the inference results for an Engagingness aspect example from the Topical-Chat dataset. As evidenced in each example, unlike G-Eval which generates scores only, MultiAgent and DEBATE enhance the quality of scores through discussions among agents. Furthermore, DEBATE, powered by the use of Devil's Advocate by the Critic, facilitates more constructive debates, thereby improving the quality of the final assessments. This comparison underscores the value of interactive evaluation methods in elevating the quality of assessment beyond simple scoring mechanisms.

\section{Examples of system messages and predefined template}
\label{system_messages}
\subsection{System messages}
The system messages corresponding to each agent used in our experiment are as follows.

\label{prompt_example}
\paragraph{Commander:}
Help me calculate the score, and tell other agents think step-by-step.
\paragraph{Scorer:}
Logically think to score the following sentence.
\paragraph{Critic(Plain):}
Do you think this score is really accurate? If you think it's not justified, please share your opinion. On the other hand, if you find the score acceptable, just say NO\_ISSUES.
\paragraph{Critic(Weakly Negative):}
Your role is to play a Devil's Advocate. Your logic has to be step-by-step. Review the score provided and assess whether the score is accurate. If there is any point to criticize, provide constructive criticism. If you find the score absolutely acceptable, just say NO\_ISSUES.
\paragraph{Critic(Moderate Negative):}
Your role is to play a Devil's Advocate. Your logic has to be step-by-step. Review the score provided and assess whether the score is accurate. Assess leniently the scores and if you think there is anything to criticize, provide feedback on the issue. If you find nothing to criticize, just say NO\_ISSUES.
\paragraph{Tie-breaker:}
You are a Tiebreaker. You will be given a news article. You will then be given one summary written for this article. You will also be given a debate log of other agents, about the rate of the summary.Your role is to take a side between the Scorer and the Critic when they disagree about the quality of a summary, and to give the final score. Scorer's role is to logically think to score the quality of a summary. Critic's role is to check if the score is justified, and give feedbacks to Scorer. You should read the Debate Log of Other Agents and decide the final score according to the following Evaluation Criteria and the Evaluation Steps.

\subsection{Predefined template (SummEval, Aspect: Coherence)}
\label{template}

\paragraph{Prompt:}\mbox{}
\noindent You will be given one summary written for a news article. Your task is to rate the summary on one metric. Please make sure you read and understand these instructions carefully. Please keep this document open while reviewing, and refer to it as needed.

\paragraph{Evaluation Criteria:}

Coherence (1-5) - the collective quality of all sentences. We align this dimension with the DUC quality question of structure and coherence whereby "the summary should be well-structured and well-organized. The summary should not just be a heap of related information, but should build from sentence to a coherent body of information about a topic."

\paragraph{Evaluation Steps:}

\begin{enumerate}
\item Read the news article carefully and identify the main topic and key points.
\item Read the summary and compare it to the news article. Check if the summary covers the main topic and key points of the news article, and if it presents them in a clear and logical order.
\item Assign a score for coherence on a scale of 1 to 5, where 1 is the lowest and 5 is the highest based on the Evaluation Criteria.
\end{enumerate}

\paragraph{Example}
\paragraph{Source Text:}
\paragraph{Summary:}
\paragraph{Evaluation Form (scores ONLY):}
\begin{itemize}
    \item Coherence:
\end{itemize}

\newpage
\onecolumn
\section{Examples of qualitative results}
\subsection{Dataset: SummEval, Aspect: Fluency}
\label{summeval_ex}
\textbf{Human Score}: Score 5

\noindent \textbf{G-Eval}: Score 1

\noindent \textbf{MultiAgent}: Score 2

\begin{itemize}
    \item \textbf{Scorer}: Upon evaluating the summary for fluency, which includes grammar, spelling, punctuation, word choice, and sentence structure, the following observations are made:
    
    1. The use of lowercase for proper nouns such as "paul merson," "andros townsend," and "tottenham" is incorrect and should be capitalized.
    
    2. The phrase "his team 's" contains an unnecessary space before the possessive apostrophe.
    
    Therefore, the score of 1 for fluency seems justified.
    
    \item \textbf{Critic}: Given these points, the low fluency score reflects significant issues with the text's adherence to standard English grammatical conventions. However, it's important to note that the scoring could also consider the overall readability and understanding of the text beyond these errors. The summary, despite its flaws, conveys the main idea of a disagreement between Paul Merson and Andros Townsend related to a soccer match and Townsend's performance. Therefore, while the identified errors justify a deduction in the fluency score, whether the score should be as low as 1 might be debatable.

    \item \textbf{Scorer}: While there are notable errors, such as the incorrect capitalization of proper nouns and the spacing issue before the possessive apostrophe, the summary does not seem to contain significant spelling mistakes or grammatical errors that render it incomprehensible. The punctuation error identified impacts readability but does not significantly hinder understanding. The rest of the sentence structure appears to be intact, allowing for the main ideas to be followed. Considering these points, the fluency score should reflect the balance between the grammatical issues and the ability of the text to convey its message. Since the text is still comprehensible and the main points are clear, a score of 2 might be more appropriate.
    
    \item \textbf{Critic}: NO ISSUE.
\end{itemize}

\noindent \textbf{DEBATE}: Score 5
\begin{itemize}
    \item \textbf{Critic}: (\textit{in response to Scorer}): While the given score of 1 for fluency based on the identified issues is understandable, it might be overly harsh when considering the overall readability and understanding conveyed by the summary. Here are several points to consider in criticizing the fluency score:

Contextual Understanding: Despite the minor errors, the overall message of the summary is clear and understandable. The reader can easily grasp the situation between Paul Merson and Andros Townsend, including the context of Townsend's late entry into a game and Merson's critical comments. The ability of the text to convey its intended message should be a significant factor in assessing fluency.

Minor Errors: The errors identified, such as the incorrect use of lowercase for proper nouns and the minor spacing issue before the possessive apostrophe, are relatively minor and do not significantly impede understanding. In the broader context of fluency, which encompasses grammar, spelling, punctuation, word choice, and sentence structure, these issues are relatively minor and could be seen as typographical or stylistic rather than fundamental flaws in fluency.

Informal Context: The summary includes a tweet ("'Paul Merson...\# 7minutes,' Merson put on Twitter."). Twitter and social media platforms often feature informal language and stylistic choices that might not adhere strictly to standard grammatical rules. The context of the content being discussed (a tweet) could justify some of the informalities or deviations from standard writing conventions.

    \item \textbf{Scorer}: Reevaluating the fluency of the summary with a more lenient perspective and considering the points raised, we can justify a higher score by emphasizing the strengths and contextual factors that mitigate the impact of the identified errors:

Overall Clarity and Coherence: Despite the lowercase usage for proper nouns and the minor spacing issue, the summary remains clear and coherent. It effectively communicates the situation involving Paul Merson and Andros Townsend, including the context of Townsend's game participation and Merson's critical commentary. This clarity in conveying the main points is a critical aspect of fluency.

Contextual Informality: The summary includes content from social media (Twitter), where informal language and unconventional grammatical choices are more acceptable. This context might justify some deviations from standard writing conventions, reflecting the original tone and style of the communication being reported.

Minor Nature of Errors: The errors present are relatively minor and do not significantly hinder comprehension. The incorrect capitalization and spacing issue are easily overlooked in casual reading and do not obscure the meaning of the text. In the broader context of fluency, which encompasses the text's ability to convey meaning effectively, these issues may not warrant a substantial penalty.

Given these considerations, while a perfect score 5 might traditionally imply an absence of errors, in this context, we can argue for a more holistic approach to fluency.

    \item \textbf{Critic}: NO ISSUE.
\end{itemize}

\subsection{Dataset: Topical-Chat, Aspect: Engagingness}
\label{topical_ex}
\textbf{Human Score}: Score 3\\
\noindent \textbf{G-Eval}: Score 1\\
\noindent \textbf{MultiAgent}: Score 2
\begin{itemize}
    \item \textbf{Scorer}: The given text is quite generic and doesn't delve into the intriguing aspect of the provided fact. Although it mentions the unique appointment by the UN, it fails to expand on this interesting detail or explain its significance, which could have made the conversation more engaging. Simply asking if the other person believes in aliens without connecting it meaningfully to the fact about the UN's appointment of ambassadors to extraterrestrials makes the response feel disconnected and dull. It misses an opportunity to spark curiosity or further discussion about the UN's actions, reasons behind such an appointment, or the implications of preparing for extraterrestrial contact. Therefore, the score of 1 for engagingness seems justified.
    
    \item \textbf{Critic}: The response could be seen as lacking depth because it does not elaborate on the fact or its implications, missing an opportunity to make the conversation more enriching. A more detailed explanation or a question that leads to a deeper discussion about the UN's decision could have elevated the engagingness of the response. However, there are some reasons why this could potentially be reconsidered for a 2.
    
    Connection to an Interesting Fact: The response introduces the fact that the UN appointed ambassadors to extraterrestrials, which is inherently an interesting and unusual piece of information. This fact alone could pique the interest of the listener and potentially lead to a more in-depth conversation about international preparations for extraterrestrial contact, the role of such an ambassador, or the implications of acknowledging the possibility of extraterrestrial life at such a high level of governance.

    Invitation for Personal Engagement: By asking, "do you believe in aliens?", the response directly engages the other person, inviting them to share their thoughts or beliefs on the topic of extraterrestrials. This personal engagement can be a critical factor in keeping a conversation interesting and dynamic.

    \item \textbf{Scorer}: Given the consideration that the response introduces an inherently interesting topic and invites personal engagement by asking for the other person's belief in aliens, coupled with the potential to spark further conversation about the UN's actions, a recalculated score could be 2. This score acknowledges that while the response might lack depth and detail, it still introduces an intriguing fact and encourages a personal exchange, making the conversation somewhat interesting and potentially engaging.
    
    \item \textbf{Critic}: NO ISSUE.
\end{itemize}

\noindent \textbf{DEBATE}: Score 3
\begin{itemize}
    \item \textbf{Critic}: (\textit{in response to Scorer}): The critique of the given score seems overly harsh when considering the potential engagement the response could generate. Asking about belief in aliens, coupled with the mention of the UN appointing ambassadors to extraterrestrials, could serve as a compelling conversation starter. This approach might intrigue the other person by blending a personal question ("do you believe in aliens?") with a surprising fact (the UN's appointment), prompting curiosity and discussion. The response subtly invites the other person to reflect on their views about extraterrestrial life and consider the international community's stance on such a possibility. It opens the door to a deeper conversation about why the UN would take such a step, what it signifies about our expectations of encountering alien life, and how seriously this topic is taken at global levels. 

    \item \textbf{Scorer}: Based on the reconsideration of the response's potential to engage by prompting curiosity and discussion about the belief in aliens and the intriguing fact of the UN appointing ambassadors for extraterrestrial contact, the response could be rated more favorably. It creates an opportunity for a deeper conversation on a topic that combines personal beliefs with international actions towards the unknown, which is inherently engaging. Therefore, a score of 3 for engagingness seems more appropriate, as the response can lead to a thought-provoking and interesting discussion.

    \item \textbf{Critic}: NO ISSUE.
\end{itemize}

\end{document}